\documentclass[10pt,a4paper]{article}

\usepackage[T1]{fontenc}
\usepackage[utf8]{inputenc}
\IfFileExists{XCharter.sty}{\usepackage{XCharter}}{\usepackage{charter}}
\usepackage{microtype}
\usepackage[
  a4paper,
  left=22mm,right=22mm,
  top=29mm,bottom=32mm,
  headheight=12pt,headsep=16pt,
  footskip=30pt
]{geometry}
\usepackage{setspace}
\usepackage{amsmath,amssymb}
\usepackage{graphicx}
\usepackage{booktabs,tabularx}
\usepackage{enumitem}
\usepackage{xcolor}
\usepackage{amsmath}

\usepackage{mathtools}
\usepackage{makecell}
\usepackage{wrapfig}
\usepackage{adjustbox}
\usepackage{siunitx}
\usepackage[ruled,vlined]{algorithm2e}
\usepackage{algpseudocode}
\usepackage{amssymb}
\usepackage{array}
\usepackage{wrapstuff}
\usepackage{colortbl}
\usepackage{arydshln}

\newcommand{\val}[2]{#1\,{\scriptsize\textcolor{gray!95}{$\pm$\,#2}}}

\newcommand{\subrule}[1]{\noalign{\vskip\aboverulesep}%
  \arrayrulecolor{black!45}\cdashline{#1}[2pt/2pt]\arrayrulecolor{black}%
  \noalign{\vskip\belowrulesep}}
\newsavebox{\algbox}
\newlength{\algheight}

\usepackage{pifont}
\let\oldding\ding
\renewcommand{\ding}[2][1]{\scalebox{#1}{\oldding{#2}}}

\definecolor{accent}{HTML}{76B900}
\definecolor{urlblue}{HTML}{000080}
\definecolor{rulegray}{gray}{0.50}
\definecolor{lightblue}{HTML}{3c78d8}
\definecolor{placeholderbg}{HTML}{F5F6F3}
\definecolor{placeholdertext}{HTML}{687260}
\usepackage{caption}
\usepackage{subcaption}
\setlist{leftmargin=*,topsep=4pt,itemsep=2pt,parsep=0pt}
\usepackage[numbers,sort&compress,round]{natbib}
\setcitestyle{numbers,round,semicolon}
\usepackage[colorlinks=true,linkcolor=accent,citecolor=accent,
  urlcolor=urlblue,bookmarksnumbered=true]{hyperref}
\newcommand{\papertitle}{Test-Time Training as Residual Memory for Robot Policies}
\newcommand{\shorttitle}{Test-Time Training as Residual Memory for Robot Policies}
\newcommand{\paperauthors}{%
  Haoxuan Wang\textsuperscript{1},
  Gengyu Zhang\textsuperscript{1},
  Ramana Rao Kompella\textsuperscript{2},
  Gaowen Liu\textsuperscript{2},
  Yan Yan\textsuperscript{1}%
}
\newcommand{\paperaffiliations}{%
  \textsuperscript{1}University of Illinois Chicago
  \quad
  \textsuperscript{2}Cisco Research
}
\newcommand{\paperurl}{https://hatchetproject.github.io/tttrm/}
\newcommand{\paperurltext}{Project Page}
\newcommand{\covernote}{}
\hypersetup{pdftitle={Research Paper Template},pdfauthor={}}

\usepackage{fancyhdr}
\renewcommand{\headrule}{\hbox to\headwidth{\color{rulegray}\leaders\hrule height 1pt\hfill}}
\renewcommand{\footrule}{\hbox to\headwidth{\color{rulegray}\leaders\hrule height 1pt\hfill}}

\fancypagestyle{cover}{%
  \fancyhf{}
  \renewcommand{\headrule}{}%
  \fancyfoot[L]{\fontsize{8}{10}\selectfont\covernote}
}
\usepackage{titlesec}
\titleformat{\section}{\fontsize{13}{16}\selectfont\bfseries}{\thesection.}{0.65em}{}
\titleformat{\subsection}{\fontsize{12}{15}\selectfont\bfseries}{\thesubsection.}{0.65em}{}
\titleformat{\subsubsection}{\normalsize\bfseries}{\thesubsubsection.}{0.65em}{}
\titleformat{\paragraph}[runin]{\normalsize\bfseries}{}{0pt}{}
\titlespacing*{\section}{0pt}{18pt plus 2pt minus 2pt}{8pt}
\titlespacing*{\subsection}{0pt}{12pt plus 2pt minus 2pt}{6pt}
\titlespacing*{\subsubsection}{0pt}{10pt}{4pt}
\titlespacing*{\paragraph}{0pt}{6pt}{0.9em}
\renewenvironment{abstract}{%
  \begingroup
  \setlength{\parskip}{0pt}%
  \titlespacing*{\section}{0pt}{14pt}{4pt}%
  \section*{Abstract}\setstretch{1.2}\bfseries\noindent
}{\par\endgroup}
\newcommand{\keywords}[1]{%
  \par\noindent{\fontsize{11}{13}\selectfont\itshape Keywords: #1}\par
}

\newcommand{\figureplaceholder}[2]{%
  \begingroup
  \setlength{\fboxsep}{0pt}%
  \colorbox{placeholderbg}{%
    \parbox[c][#1][c]{\linewidth}{%
      \centering\color{placeholdertext}\normalfont\small #2\par
    }%
  }%
  \endgroup
}

\newcommand{\makepaperheading}{%
  \begingroup
  \setstretch{1}%
  \setlength{\parindent}{0pt}%
  \setlength{\parskip}{0pt}%
  \raggedright
  {\color{black}\fontsize{17}{21}\selectfont\bfseries\papertitle\par}%
  \vspace{12pt}%
  {\fontsize{9}{12}\selectfont\bfseries\paperauthors\par}%
  \vspace{3pt}%
  {\fontsize{8}{10}\selectfont\paperaffiliations\par}%
  \vspace{3pt}%
  {\fontsize{8}{10}\selectfont
    \href{\paperurl}{\texttt{\paperurltext}}\par}%
  \endgroup
}

\newif\ifcoverpage
\coverpagefalse

\begin{document}

\ifcoverpage
\else
  \thispagestyle{cover}
  \makepaperheading
\fi

\begin{abstract}
Memory is essential for long-horizon robotic manipulation, where successful actions may depend on past events that are no longer recoverable from the current observation. As episodes grow longer, however, retaining the full history becomes increasingly costly, creating a fundamental scalability challenge for memory-augmented policies.
Existing approaches address this challenge by either storing selected past observations in a bounded memory bank or compressing interaction history into a fixed-size parametric state through Test-Time Training (TTT).
Yet these formulations do not explicitly distinguish between historical information that can already be recovered from the policy's current context and information that must persist beyond it.
We introduce \textbf{TTT-RM}, which repurposes TTT as \emph{Residual Memory}, using fast weights not to generically compress history but to complement a bounded memory bank by preserving task-relevant historical information that cannot be recovered from the policy's current context.
Concretely, TTT-RM learns a history decoder that reconstructs historical representations from the current observation and retrieved memory. 
The resulting reconstruction residual captures what this context fails to explain and serves as the learning target for TTT. The TTT slow weights are optimized against this residual target so that online fast-weight updates learn to encode complementary historical information over time. The fast-weight state is then queried to produce a residual memory representation that conditions action generation.
Extensive experiments on memory-intensive simulation benchmarks and real-world tasks show that TTT-RM consistently improves across multiple memory designs, outperforms diverse baselines, and supports sustained execution on a three-minute, eight-stage stowing task.
\end{abstract}

\section{Introduction}
Long-horizon robotic manipulation is often partially observable, such that the current observation alone may not contain all information needed to determine the next action~\citep{partial_obs,memorybench}. An object may become occluded after being moved, an instruction may need to be recalled long after it is issued, or the next action may depend on which subgoals have already been completed and on the outcomes of earlier actions. Consequently, task-relevant information is distributed across the interaction history, requiring the policy to integrate past evidence to recover the latent task state. Effective policies therefore require memory mechanisms that preserve relevant historical information for history-dependent action prediction.

Yet many state-of-the-art robot foundation models~\citep{openvla,gr00t,pi05,cosmos_policy,fastwam} condition action prediction primarily on the current observation or a limited temporal context. While sufficient for local visuomotor reasoning, this limited context becomes inadequate when task-relevant information originates much earlier in the episode. Recent work addresses this limitation through two prominent memory designs. The first store selected multimodal observations as individually retrievable records in a bounded memory bank~\citep{memer,memoryvla,memorywam}, which we collectively refer to as \textbf{Explicit Memory (EM)}. EM preserves event-specific information and allows the policy to revisit past observations when needed. Its coverage, however, is constrained by the bank capacity, so relevant events may be omitted or evicted as the interaction history grows. Increasing the bank capacity can improve coverage but incurs additional storage and retrieval costs.

\textbf{Test-Time Training (TTT)} offers an alternative memory mechanism by encoding interaction history into fast weights that are updated online as new observations arrive~\citep{robottt,wamttt}. Unlike EM, which allocates storage to individually retained records, TTT continually integrates history into this fixed-size memory state. This fixed memory footprint avoids increasing storage with the amount of retained history, but shifts the challenge to determining what information should occupy its limited representational capacity. EM and TTT therefore provide complementary forms of persistence, with the former preserving individually retrievable events and the latter compressing history into its internal weights. Given TTT's limited representational capacity, our key insight is to specialize its fast weights for task-relevant historical information that remains unrecoverable from the policy's existing context, rather than using them as a generic history compressor.

We therefore introduce \textbf{TTT-RM}, a framework that augments robot policies equipped with Explicit Memory by repurposing TTT as \textbf{Residual Memory} to preserve task-relevant history that cannot be recovered from the policy's current observation and retrieved memory. During training, a history decoder reconstructs historical representations from this context, and the resulting reconstruction residual captures what remains unexplained. This residual supervises the TTT slow weights so that online fast-weight updates learn to encode complementary historical information over the course of an episode. The fast-weight state is then queried to produce a Residual Memory representation, which is combined with retrieved memory to condition action generation. TTT-RM reuses the visual stream already processed by the Explicit Memory module, requiring no additional encoder or VLM forward passes. At inference time, the history decoder is discarded, while the fast-weight state is updated at every policy call and reset at the start of each episode.

TTT-RM is compatible with multiple Explicit Memory designs, making it a modular extension to existing history-dependent robot policies. Across multiple memory-intensive simulation benchmarks, TTT-RM consistently outperforms diverse baselines, demonstrating robust gains across different task settings. On real-world manipulation tasks, it improves task progress by 10\% to 26.5\% and supports sustained execution on a three-minute, eight-stage long-horizon stowing task.

\begin{figure}[t]
    \centering
    \includegraphics[width=1.0\linewidth, trim=0 200 20 130, clip]{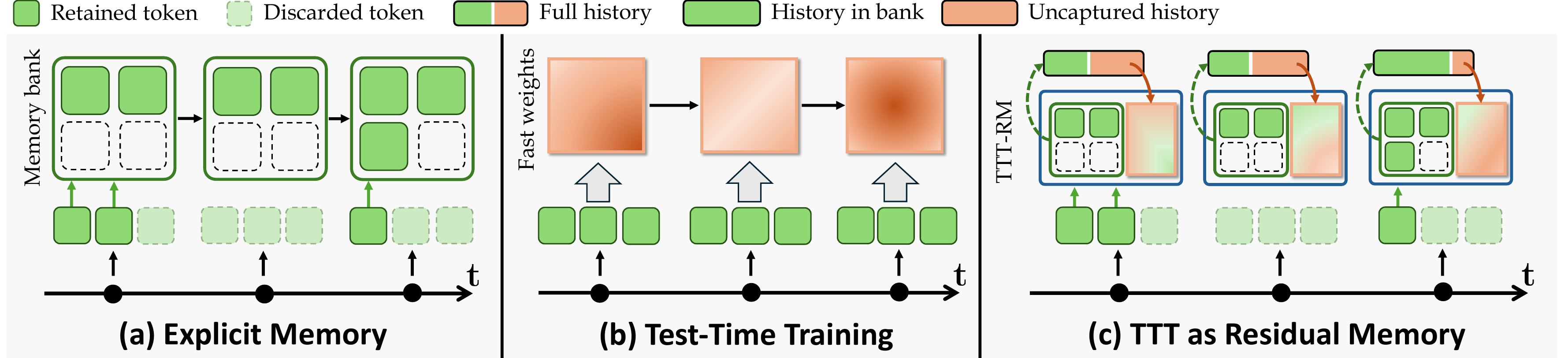}
    \vspace{-12pt}
    \captionsetup{
        justification=raggedright,
        singlelinecheck=false
    }
    \caption{\textbf{Comparison of memory paradigms.} Incoming observations are represented as tokens along the timeline. \textbf{(a)} \textbf{Explicit Memory} retains selected historical information in a bounded memory bank, giving direct access to only a sparse subset of the preceding history.
    \textbf{(b)} \textbf{Test-Time Training} instead compresses history into a fixed-size fast-weight state whose size does not grow with episode length.
    \textbf{(c)} \textbf{TTT as Residual Memory.} TTT is repurposed to preserve only the historical information not recoverable from the current observation and retrieved memory.}
    \label{fig:compare}
\end{figure}

\section{Related Work}
\subsection{Explicit Memory for Robot Learning}
A growing line of work augments robot policies with Explicit Memory, storing selected history as retrievable records. Existing methods differ in what they retain and how they organize it. MemER~\citep{memer} retains task-relevant keyframes from prior robot experience. MemoryVLA~\citep{memoryvla} instead maintains a perceptual-cognitive memory bank spanning visual details and higher-level semantic information. EventVLA~\citep{eventvla} stores visual anchors and sparse task-critical events, whereas Mem-0~\citep{rmbench} retains keyframes around subtask completion to support memory-conditioned planning and execution.
Other approaches organize history across representations or temporal scales. MemoryWAM~\citep{memorywam} integrates recent frames, event-boundary anchors, and compact gist tokens. MEM~\citep{mem} combines video-based short-term memory with text-based long-term memory, and DiM-WAM~\citep{dim_wam} maintains bounded banks of complementary historical event tokens. LaMem-VLA~\citep{lamem_vla} accesses separate short- and long-term memory stores before encoding selected evidence into latent memory tokens. Together, these methods extend the policy's effective temporal context through explicit storage and retrieval, but the information available at each decision point remains bounded by the memory budget and the records selected for retention.

\subsection{Test-Time Training and Fast-Weight Memory}
Test-Time Training (TTT) was originally introduced as an adaptation paradigm in which model parameters are optimized on unlabeled test samples using a self-supervised objective~\citep{ttt_0}. Building on the broader idea of fast-weight models~\citep{fast_weight,fast_weight_2,fast_weight_3}, which encode past information through dynamically updated parameters, recent work has extended TTT to sequence modeling by treating fast weights as a recurrent memory state updated online as new inputs arrive~\citep{ttt_rnn,titans,atlas}.
Beyond language modeling, related test-time optimization has also been explored for novel view synthesis and autoregressive video generation~\citep{ttt_done_right}. In robot learning, TTT has recently been adopted as a memory mechanism for history-dependent control. RoboTTT~\citep{robottt} integrates TTT layers into robot policies to compress interaction history into recurrent fast-weight states. WAM-TTT~\citep{wamttt} instead meta-trains an adaptive memory that is updated from unlabeled human videos at deployment. Our work builds on this memory interpretation of TTT, but fundamentally changes what the fast-weight state is trained to preserve. Rather than generically compressing interaction history, TTT-RM specializes the fast-weight state to preserve residual information that is not recoverable from the current observation and retrieved memory.

\section{Preliminaries}
\textbf{History-dependent robot policies.}
Let $\ell$ denote a language instruction and $\mathbf{o}_t$ the multimodal observation at policy step $t$, including visual and proprioceptive inputs. 
Under partial observability, the current observation $\mathbf{o}_t$ may be insufficient for action prediction, since similar observations can arise from different histories and require different subsequent actions.
We therefore consider policies equipped with an episode-scoped memory state $\mathsf{S}_t$ that preserves information from preceding interactions. The policy predicts
\begin{equation}
    \mathbf{A}_t
    \sim
    \pi_\theta\!\left(
        \,\cdot \mid
        \ell,\mathbf{o}_t,\mathsf{S}_t
    \right),
\end{equation}
where $\theta$ denotes the slow policy parameters that remain fixed within an episode. The memory state $\mathsf{S}_t$ is updated causally across policy calls and reset at the beginning of each episode. We next consider two mechanisms for maintaining $\mathsf{S}_t$: Explicit Memory and Test-Time Training.

\noindent\textbf{Explicit Memory.}
We use Explicit Memory (EM) to denote mechanisms that retain a bounded set of historical representations for direct access by the policy.
Let $\mathbf{z}_{s,p}$ denote a visual feature extracted at policy step $s$ and spatial location $p$, and let $\rho(s,p)$ be its spatiotemporal encoding.
At policy step $t$, the EM mechanism maintains a bounded index set $\mathcal{I}_t$ and the corresponding memory
\begin{equation}
\begin{aligned}
\mathcal{I}_t &\subseteq
\left\{
(s,p) \,\middle|\, 1 \leq s < t,\; p \in \mathcal{P}_s
\right\},
\qquad
|\mathcal{I}_t| \leq K, \\
\mathcal{M}_t &=
\left\{
\bigl(\mathbf{z}_{s,p}, \rho(s,p)\bigr)
\,\middle|\,
(s,p) \in \mathcal{I}_t
\right\},
\end{aligned}
\label{eq:eim}
\end{equation}
where $\mathcal{P}_s$ denotes the set of spatial locations at step $s$, and $K$ denotes the memory budget.
For an EM-based policy, the generic memory state specializes to
$\mathsf{S}_t = \mathcal{M}_t$, giving the policy direct access to the retained historical representations.

\noindent\textbf{Test-Time Training formulation.} 
In contrast to EM, parametric memory encodes interaction history into a fixed-size internal state~\citep{robomamba,mtil,robottt}. Among such approaches, we focus on Test-Time Training, which realizes this state as fast weights updated online through an inner-loop learning objective~\citep{ttt_rnn}. Crucially, this objective provides an explicit learning signal for determining what information is written into the fast-weight state.
Let $f_{\boldsymbol{\phi}_t}$ denote the TTT memory model with fast weights $\boldsymbol{\phi}_t$.
Given an input representation $\mathbf{x}_t$, learned mappings $\mathcal{K}$, $\mathcal{V}$, and $\mathcal{Q}$ produce the corresponding key, value, and query,
\begin{equation}
\mathbf{k}_t = \mathcal{K}(\mathbf{x}_t), \qquad
\mathbf{v}_t = \mathcal{V}(\mathbf{x}_t), \qquad
\mathbf{q}_t = \mathcal{Q}(\mathbf{x}_t).
\end{equation}
TTT then performs a \emph{write} followed by a \emph{read}.
The \emph{write} operation updates the fast weights by associating the key with its target value through the inner-loop objective
\begin{align}
\mathcal{L}^{\mathrm{fast}}_t(\boldsymbol{\phi})
&=
\left\|
f_{\boldsymbol{\phi}}(\mathbf{k}_t)-\mathbf{v}_t
\right\|_2^2, \\
\boldsymbol{\phi}_t
&=
\boldsymbol{\phi}_{t-1}
-
\eta_t
\left.
\nabla_{\boldsymbol{\phi}}
\mathcal{L}^{\mathrm{fast}}_t(\boldsymbol{\phi})
\right|_{\boldsymbol{\phi}=\boldsymbol{\phi}_{t-1}},
\label{eq:ttt_inner}
\end{align}
where $\eta_t$ denotes the inner-loop learning rate. 
The \emph{read} operation evaluates the updated fast-weight model at the query, $f_{\boldsymbol{\phi}_t}(\mathbf{q}_t)$, without further modifying the fast weights. The slow parameters, including those of $\mathcal{K}$, $\mathcal{V}$, $\mathcal{Q}$, and the policy $\theta$, are optimized across training episodes by the outer objective and remain fixed during deployment. In contrast, the fast weights are initialized at the start of each episode and updated at every policy call. 
For TTT, the memory state entering policy step $t$ is $\mathsf{S}_t=\boldsymbol{\phi}_{t-1}$, which is updated to $\boldsymbol{\phi}_t$ after processing the current input. Because the dimensionality of the fast-weight state is independent of episode length, TTT can continually integrate history within a fixed memory footprint. The effectiveness of this fixed-capacity memory therefore depends critically on what information the fast-weight state is trained to preserve.

\section{Methodology}
\begin{figure}[t]
\centering
\begin{lrbox}{\algbox}
\begin{minipage}{0.46\linewidth}
\begin{algorithm}[H]
\caption{\textsc{TTT-RM} Training}
\label{alg:ttt_rm}
\KwIn{Policy with EM, history decoder $\mathcal{D}$, TTT fast weights $\boldsymbol{\phi}$}
\tcp{\textcolor{lightblue}{History Decoder Learning}}
Train $\mathcal{D}$ with Eq.~\ref{eq:decoder_training} and freeze;

\tcp{\textcolor{lightblue}{Residual Memory Learning}}
\For{each training stage and trajectory}{
    Initialize $\boldsymbol{\phi} \leftarrow \boldsymbol{\phi}_0$\;
    \For{policy step $t$}{
        Retain tokens from $\mathbf{o}_t$ into $\mathcal{M}_t$\;
        Update $\boldsymbol{\phi}$ from visual write pairs
        (Eqs.~\ref{eq:ttt_inner}, \ref{eq:inner_loss})\;
        Construct $\mathbf{r}_{t;s,p}$ and read $\boldsymbol{\delta}_{t;s,p}$
        (Eqs.~\ref{eq:fast_residual}, \ref{eq:residual_target})\;
        Optimize slow weights with $\mathcal{L}_t$
        (Eq.~\ref{eq:final_loss})\;
    }
}
\end{algorithm}
\end{minipage}
\end{lrbox}
\setlength{\algheight}{\dimexpr\ht\algbox+\dp\algbox\relax}
\begin{minipage}[t]{0.58\linewidth}
  \vspace{0pt}
  \centering
  \includegraphics[height=\algheight, width=\linewidth, keepaspectratio,
                   trim=300 160 430 110, clip]{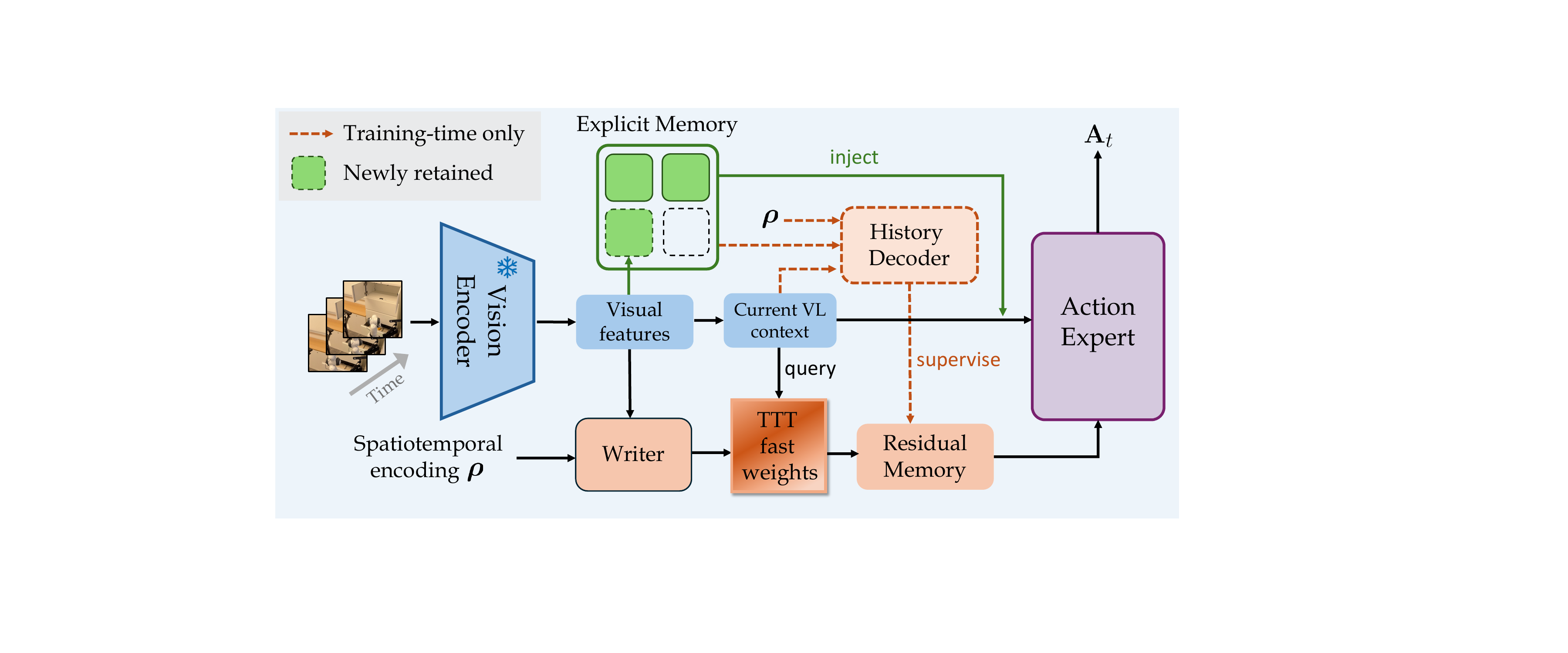}\par
  \vspace{-4pt}
  \captionsetup{
        justification=raggedright,
        singlelinecheck=false
    }
  \captionof{figure}{\textbf{Overview of TTT-RM.} Frozen visual features are selectively stored in Explicit Memory and written into TTT fast weights through the visual-stream writer. During training, the history decoder estimates what past information is recoverable from the existing policy context, and its reconstruction residual supervises Residual Memory. At inference, the decoder is removed, while Explicit Memory and Residual Memory jointly condition action generation.}
  \label{fig:pipeline}
\end{minipage}%
\hfill
\begin{minipage}[t]{0.40\linewidth}
  \vspace{0pt}
  \usebox{\algbox}
\end{minipage}
\vspace{-0pt}
\end{figure}

\textbf{Overview.}
We consider robot policies equipped with Explicit Memory and seek to use TTT as a complementary memory substrate rather than a generic history compressor. To this end, we introduce \textbf{TTT-RM}, which repurposes TTT as Residual Memory. Fig.~\ref{fig:pipeline} illustrates the overall pipeline. TTT-RM first learns a history decoder that reconstructs historical information from the policy's current observation and retrieved memory. The resulting reconstruction residual captures the history unexplained by this context and supervises TTT so that online fast-weight updates learn to encode it over time. The fast-weight state is then queried to produce a Residual Memory representation, which is combined with the retrieved memory to condition action generation. Secs.~\ref{sec:decoder_learning}--\ref{sec:implementation} detail the history decoder, residual-memory learning, and training and inference procedures.

\subsection{History Decoder Learning}
\label{sec:decoder_learning}
Under a finite memory budget, Explicit Memory retains only a subset of historical representations. Yet historical information is often redundant, and representations not retained by EM may still be recoverable from the current observation and retrieved memory. Treating all unretained history as missing would therefore overestimate what additional memory needs to preserve.
We therefore define the target of Residual Memory by \emph{what remains unexplained} by the policy's existing context, rather than by what EM fails to retain. We operationalize this distinction with a history decoder that reconstructs historical representations from the current observation and retrieved memory.

Given a visual feature $\mathbf{z}_{s,p}$ extracted by the frozen encoder at policy step $s$ and spatial location $p$, we define the corresponding history target as
\begin{equation} 
\mathbf{y}_{s,p} = \mathbf{P}\mathbf{z}_{s,p}, 
\label{eq:history_target}
\end{equation} 
where $\mathbf{P}$ is a fixed semi-orthogonal projection into the representation space of the TTT module. Because both the visual encoder and \(\mathbf{P}\) remain fixed during training, \(\mathbf{y}_{s,p}\) provides a stationary feature-space target for the visual content at \((s,p)\).
To estimate which historical representations are already recoverable from the policy's available context, we train a fixed-capacity history decoder $\mathcal{D}$. Given the current observation, retrieved memory, and language instruction, the decoder predicts the history target at any historical address $(s,p)$, with $s < t$: 
\begin{equation}
\hat{\mathbf{y}}_{t;s,p} = \mathcal{D}\bigl(\mathcal{M}_t,\mathbf{o}_t,\ell,\rho(s,p),t-s\bigr).
\label{eq:decoder_prediction}
\end{equation}
Here, $\rho(s,p)$ sinusoidally encodes the target's spatiotemporal address, while $t-s$ separately encodes its age relative to the current policy step.

To train the decoder across historical targets, let $\mathcal{C}_t \subseteq \left\{ (s,p) \,\middle|\, 1 \leq s < t,\; p \in \mathcal{P}_s \right\}$ denote the set of historical addresses used for reconstruction at policy step $t$. We train $\mathcal{D}$ by minimizing
\begin{equation}
\mathcal{L}^{\mathrm{dec}}_t = \frac{1}{|\mathcal{C}_t|}\sum_{(s,p)\in\mathcal{C}_t}\left\|\hat{\mathbf{y}}_{t;s,p} - \operatorname{sg}\!\left(\mathbf{y}_{s,p}\right) \right\|_2^2,
\label{eq:decoder_training}
\end{equation}
where $\operatorname{sg}(\cdot)$ denotes stop-gradient operator.
After decoder training, $\mathcal{D}$ is frozen and used as a fixed reference throughout subsequent policy training.

\subsection{Test-Time Training as Residual Memory}
\label{sec:ttt_residual}
With the history decoder fixed, we next repurpose TTT as \emph{Residual Memory}. TTT-RM consists of three key elements. First, an inner-loop update writes incoming visual information into the fast-weight state. Second, the updated state is queried to produce a Residual Memory representation for action generation. Third, an outer-loop objective uses supervision from the frozen history decoder to optimize the slow parameters, shaping the online updates to preserve historical information that is not recoverable from the current observation and retrieved memory.

\noindent\textbf{Inner-loop fast weight update.}
We first specify how incoming visual information is written into the TTT fast-weight state. Unlike prior TTT-based robot policies that derive write pairs from intermediate action-expert representations~\citep{robottt}, TTT-RM constructs them directly from visual features produced by the frozen encoder. At policy step $t$, the visual feature $\mathbf{z}_{t,p}$ at spatial location $p$ and its spatiotemporal encoding $\rho(t,p)$ are mapped to a key, while the corresponding history target serves as the value:
\begin{equation}
\mathbf{k}_{t,p}
=
\mathcal{K}
\left(
\operatorname{sg}\!\left(\mathbf{z}_{t,p}\right),
\rho(t,p)
\right),
\qquad
\mathbf{v}_{t,p}
=
\mathbf{y}_{t,p},
\label{eq:ttt_rs_fast}
\end{equation}
where $\mathcal{K}$ is the learnable key mapping and $\mathbf{y}_{t,p}$ is the fixed history target defined in Eq.~\ref{eq:history_target}.
We apply the inner-loop objective over all spatial locations $p\in\mathcal{P}_t$, and and parameterize the write target in residual form for training stability:
\begin{equation}
\mathcal{L}^{\mathrm{fast}}_t(\boldsymbol{\phi}) = \frac{1}{|\mathcal{P}_t|} \sum_{p \in \mathcal{P}_t} \left\| f_{\boldsymbol{\phi}}(\mathbf{k}_{t,p}) - \left( \operatorname{sg}\!\left(\mathbf{v}_{t,p}\right) - \mathbf{k}_{t,p} \right) \right\|_2^2.
\label{eq:inner_loss} 
\end{equation}
The fast weights are then updated according to Eq.~\ref{eq:ttt_inner}. 

\noindent\textbf{Residual Memory construction.} 
For memory readout, the reader constructs a query $\mathbf{q}_{t;s,p}$ from the current policy context and the spatiotemporal address $(s,p)$. Applying the updated fast-weight model to this query produces an address-conditioned readout $f_{\boldsymbol{\phi}_t}(\mathbf{q}_{t;s,p})$. To isolate the contribution of within-episode fast-weight updates, we subtract the corresponding readout under the episode-initial state and define the Residual Memory representation as
\begin{equation} 
\boldsymbol{\delta}_{t;s,p} = f_{\boldsymbol{\phi}_t}(\mathbf{q}_{t;s,p}) - f_{\boldsymbol{\phi}_0}(\mathbf{q}_{t;s,p}),
\label{eq:fast_residual}
\end{equation}
where $\boldsymbol{\phi}_0$ denotes the fast-weight state at the beginning of the episode.

\noindent\textbf{Outer-loop optimization.} 
With the history decoder frozen, we define the Residual Memory target as the portion of each historical representation unexplained by the decoder:
\begin{equation} 
\mathbf{r}_{t;s,p} = \operatorname{sg}\!\left( \mathbf{y}_{s,p} - \hat{\mathbf{y}}_{t;s,p} \right). 
\label{eq:residual_target}
\end{equation} 

We optimize the TTT slow parameters so that the Residual Memory readout approximates this decoder-defined target. Over the set of historical addresses $\mathcal{C}_t$, the residual objective is
\begin{equation}
\mathcal{L}^{\mathrm{res}}_t = \frac{1}{|\mathcal{C}_t|} \sum_{(s,p)\in\mathcal{C}_t} \left\| \boldsymbol{\delta}_{t;s,p} - \mathbf{r}_{t;s,p} \right\|_2^2. 
\label{eq:residual_loss}
\end{equation}
Through this objective, the slow write and read parameters are trained so that within-episode fast-weight dynamics preserve information not recoverable from the existing policy context.

To make this information available for control, we condition the action expert on the resulting residual readouts. Each action block accesses the same set of address-conditioned residual representations through a lightweight attention adapter over the queried spatiotemporal addresses and their corresponding readouts $\boldsymbol{\delta}_{t;s,p}$. The adapter and other trainable slow parameters are jointly optimized with the flow-matching action objective $\mathcal{L}^{\mathrm{act}}_t$. The overall outer-loop objective is
\begin{equation}
\mathcal{L}_t = \mathcal{L}^{\mathrm{act}}_t + \lambda_{\mathrm{res}} \mathcal{L}^{\mathrm{res}}_t,
\label{eq:final_loss}
\end{equation}
where $\lambda_{\mathrm{res}}$ balances Residual Memory supervision against action prediction.

\subsection{Training and Inference}
\label{sec:implementation}
\textbf{Training.} We train TTT-RM in three stages. First, we train the history decoder $\mathcal{D}$ using Eq.~\ref{eq:decoder_training} and freeze it as the reference for constructing residual targets during subsequent policy training. Second, we train only the newly introduced Residual Memory components while keeping the inherited Explicit Memory and action policy frozen. At this stage, the fast-weight initialization $\boldsymbol{\phi}_0$, visual writer $\mathcal{K}$, reader $\mathcal{Q}$, and zero-initialized action adapters are jointly optimized using Eq.~\ref{eq:final_loss}. Finally, we fine-tune the full policy with Eq.~\ref{eq:final_loss}, allowing the inherited Explicit Memory and action components to adapt jointly to Residual Memory. Alg.~\ref{alg:ttt_rm} summarizes the training procedure.

\noindent\textbf{Inference.}
At inference time, the history decoder is discarded and all slow parameters remain fixed. The fast weights are initialized from $\boldsymbol{\phi}_0$ at the beginning of each episode and updated once per policy call as new observations arrive. Within each policy call, the updated fast-weight state remains fixed across all flow-integration steps and is reused to condition action generation, avoiding repeated inner-loop updates for the same observation. The fast-weight update is implemented as an explicit gradient step within the forward pass. During training, the same update remains differentiable, allowing gradients to propagate through the inner-loop step to the slow parameters.

\begin{table}[t]
  \centering
  \renewcommand{\arraystretch}{1.15}
  \captionsetup{
        justification=raggedright,
        singlelinecheck=false
    }
  \caption{\textbf{Results on RoboMME.} Performance is measured as average success rate (\%). Each TTT-RM variant augments the corresponding Explicit Memory baseline with Residual Memory.}
  \vspace{-4pt}
  \label{tab:robomme_result}
  \begin{tabular}{lcccc@{\hspace{20pt}}c}
    \toprule
    \textbf{Method} & \texttt{Counting} & \texttt{Permanence} & \texttt{Reference} & \texttt{Imitation} & \texttt{\textbf{AVG}} \\
    \midrule
    Base~\citep{pi05} & \val{23.3}{0.7} & \val{17.1}{2.9} & \val{20.0}{2.5} & \val{13.3}{1.4} & \val{18.4}{0.8} \\
    SAM2Act+~\citep{memorybench} & \val{35.3}{1.4} & \val{26.0}{1.4} & \val{16.8}{1.4} & \val{7.3}{0.6} & \val{21.4}{0.6} \\
    MemER~\citep{memer} & \val{48.8}{2.5} & \val{53.2}{1.4} & \val{38.0}{2.1} & \val{29.5}{1.2} & \val{42.4}{0.9} \\
    \midrule
    \multicolumn{6}{l}{\textsc{Test-Time Training}~\citep{robomme}} \\
    \quad w/ \emph{Expert} & \val{22.1}{6.9} & \val{20.8}{1.9} & \val{20.0}{2.5} & \val{7.1}{0.7} & \val{17.5}{1.1} \\
    \quad w/ \emph{Modul} & \val{34.5}{3.4} & \val{22.2}{4.1} & \val{20.4}{1.8} & \val{10.7}{1.6} & \val{21.9}{1.5} \\
    \midrule
    \multicolumn{6}{l}{\textsc{Explicit Memory}~\citep{robomme}} \\
    \quad TokenDrop \ w/ \emph{Expert} & \val{57.9}{0.7} & \val{25.0}{2.2} & \val{23.8}{2.2} & \val{31.3}{1.3} & \val{34.5}{0.4} \\
    \quad FrameSamp w/ \emph{Expert} & \val{67.1}{0.7} & \val{24.6}{0.7} & \val{26.7}{0.7} & \val{28.3}{3.2} & \val{36.7}{1.0} \\
    \subrule{1-6}
    \quad TokenDrop \ w/ \emph{Modul} & \val{44.2}{6.2} & \val{19.6}{0.7} & \val{23.3}{0.7} & \val{36.7}{1.4} & \val{30.9}{1.1} \\
    \quad FrameSamp w/ \emph{Modul} & \val{60.0}{1.3} & \val{25.6}{0.6} & \val{30.6}{1.9} & \val{55.0}{3.8} & \val{42.8}{1.9} \\
    \midrule
    \multicolumn{6}{l}{\textbf{\textsc{TTT-RM (ours)}}} \\
    \quad TokenDrop \ w/ \emph{Expert} & \val{60.8}{1.4} & \val{26.3}{1.3} & \val{28.3}{0.7} & \val{32.9}{3.2} & \val{\textbf{37.1}}{1.2} \\
    \quad FrameSamp w/ \emph{Expert} & \val{69.4}{0.6} & \val{25.6}{0.6} & \val{27.5}{5.0} & \val{33.1}{1.9} & \val{\textbf{38.9}}{0.8} \\
    \subrule{1-6}
    \quad TokenDrop \ w/ \emph{Modul} & \val{50.0}{1.3} & \val{32.5}{0.8} & \val{29.4}{1.9} & \val{40.0}{1.3} & \val{\textbf{38.0}}{0.5} \\
    \quad FrameSamp w/ \emph{Modul} & \val{62.1}{0.7} & \val{30.0}{3.3} & \val{32.1}{1.4} & \val{57.5}{1.3} & \val{\textbf{45.4}}{1.0} \\
    \bottomrule
  \end{tabular}
\end{table}

\section{Experiments}
\label{sec:exp}
We evaluate TTT-RM in both simulation and real-world settings across tasks requiring diverse forms of long-horizon memory. We compare against strong memory-based baselines and conduct targeted ablations and diagnostic analyses to isolate the contribution of individual design choices.

\subsection{Simulation Benchmarks}
We first evaluate TTT-RM on two memory-intensive manipulation benchmarks, RoboMME~\citep{robomme} and RMBench~\citep{rmbench}. RoboMME is a large-scale benchmark for long-horizon, history-dependent manipulation, comprising 16 tasks grouped into four suites: \texttt{Counting}, \texttt{Permanence}, \texttt{Reference}, and \texttt{Imitation}. These suites target temporal, spatial, object, and procedural memory, respectively, covering diverse forms of history dependence. RMBench contains nine dual-arm manipulation tasks at two levels of memory complexity. \(M(1)\) tasks require a single past observation to disambiguate the current state, whereas \(M(n)\) tasks require integrating information from multiple past observations to determine the appropriate action. Together, these benchmarks cover both diverse memory types and varying degrees of historical dependence. For a controlled comparison, all methods follow the same evaluation protocol within each benchmark.

\noindent\textbf{Baselines.} We compare \textsc{TTT-RM} against the following baselines. Unless otherwise noted, baselines with learned action policies use the same $\pi_{0.5}$~\citep{pi05} backbone:
\begin{itemize}[leftmargin=10pt,topsep=-3pt,itemsep=1pt,partopsep=1pt,parsep=1pt]
    \item \textbf{Base}: The memory-free policy that conditions action prediction only on the current observation.
    \item \textbf{SAM2Act+}~\citep{memorybench}: A perceptual-memory baseline that uses SAM2~\citep{sam2} to maintain a memory bank over past observations and predicts discrete keyframe actions. The predicted actions are executed by an external motion planner rather than a learned action policy.
    \item \textbf{MemER}~\citep{memer}: A dual-system baseline that stores selected keyframe observations and uses a VLM~\citep{qwen3} to infer symbolic subgoals from the accumulated memory. The resulting subgoals are executed by the base $\pi_{0.5}$ policy.
    \item \textbf{TTT variants}: We evaluate two TTT instantiations from RoboMME, \emph{Modul} (memory-as-modulator) and \emph{Expert} (memory-as-expert). \emph{Modul} uses the memory representation to modulate intermediate action-expert activations through adaptive LayerNorm~\citep{layernorm}. \emph{Expert} instead introduces a separate memory model that interacts with the action expert during prediction. We exclude the \emph{memory-as-context} design, which directly appends memory to the input context and thereby changes both input sequence length and the VLM representations.
    \item \textbf{Explicit Memory variants}: We evaluate the two best-performing Explicit Memory mechanisms from RoboMME, \textbf{TokenDrop} and \textbf{FrameSamp}. TokenDrop retains informative spatial tokens from historical observations, whereas FrameSamp selects historical frames and retains a pooled representation for each selected frame. Each of these two memory mechanisms is evaluated with both the \emph{Modul} and \emph{Expert} architectures described above.
\end{itemize}

\noindent\textbf{Results.}
Tab.~\ref{tab:robomme_result} reports the results on RoboMME. Standalone TTT yields mixed performance across integration designs. The \emph{Modul} variant improves the average success rate from 18.4\% to 21.9\%, whereas \emph{Expert} slightly underperforms the base policy at 17.5\%. This variation suggests that simply introducing a fast-weight memory state does not consistently improve history-dependent control, and its effectiveness depends on how the resulting memory representation interacts with the action policy. By contrast, all Explicit Memory variants substantially outperform the base policy, showing the benefit of explicit retrieval for the memory demands of RoboMME under the evaluated budgets and task horizons. 
TTT-RM further improves every configuration, with average success-rate gains ranging from 2.2 to 7.1 percentage points. The gains are also consistent across task suites, with TTT-RM outperforming the corresponding EM baseline in all 16 configuration--suite comparisons. The best-performing configuration, FrameSamp w/ \emph{Modul}, improves from 42.8\% to 45.4\%. Overall, standalone TTT performance depends on how fast-weight memory is integrated into the policy, whereas TTT-RM provides consistent gains across different Explicit Memory mechanisms, integration designs, and forms of history dependence.

\begin{wraptable}{r}{0.5\linewidth}
\vspace{-14pt}
\centering
\captionsetup{
    justification=raggedright,
    singlelinecheck=false
}
\caption{\textbf{Results on RMBench.}
Performance is measured as average success rate (\%).}
\label{tab:rmbench}
\vspace{-4pt}
\footnotesize
\setlength{\tabcolsep}{3.5pt}
\begin{adjustbox}{width=\linewidth}
\begin{tabular}{@{}lrrrr@{}}
\toprule
\textbf{Method}
& \multicolumn{1}{c}{Base}
& \multicolumn{1}{c}{TTT}
& \multicolumn{1}{c}{EM}
& \multicolumn{1}{c@{}}{\textbf{Ours}} \\
\midrule
$M(1)$
& \val{15.6}{1.7}
& \val{11.2}{0.7}
& \val{30.5}{3.3}
& \val{\textbf{36.5}}{4.1} \\
$M(n)$
& \val{8.8}{0.8}
& \val{10.3}{0.8}
& \val{10.8}{0.3}
& \val{\textbf{15.8}}{0.3} \\
\bottomrule
\end{tabular}
\end{adjustbox}
\vspace{-12pt}
\end{wraptable}

Tab.~\ref{tab:rmbench} reports RMBench results using an Explicit Memory configuration analogous to FrameSamp w/ \emph{Modul}. Standalone TTT yields mixed performance, underperforming the base policy on \(M(1)\) but modestly improving over it on \(M(n)\), while EM is the strongest baseline in both settings. TTT-RM further improves over EM by 6\% on \(M(1)\) and 5\% on \(M(n)\).

\begin{figure}[t]
    \centering
    \adjustbox{width=1.0\linewidth}{
    \includegraphics[width=1.0\linewidth, trim=10 150 50 130, clip]{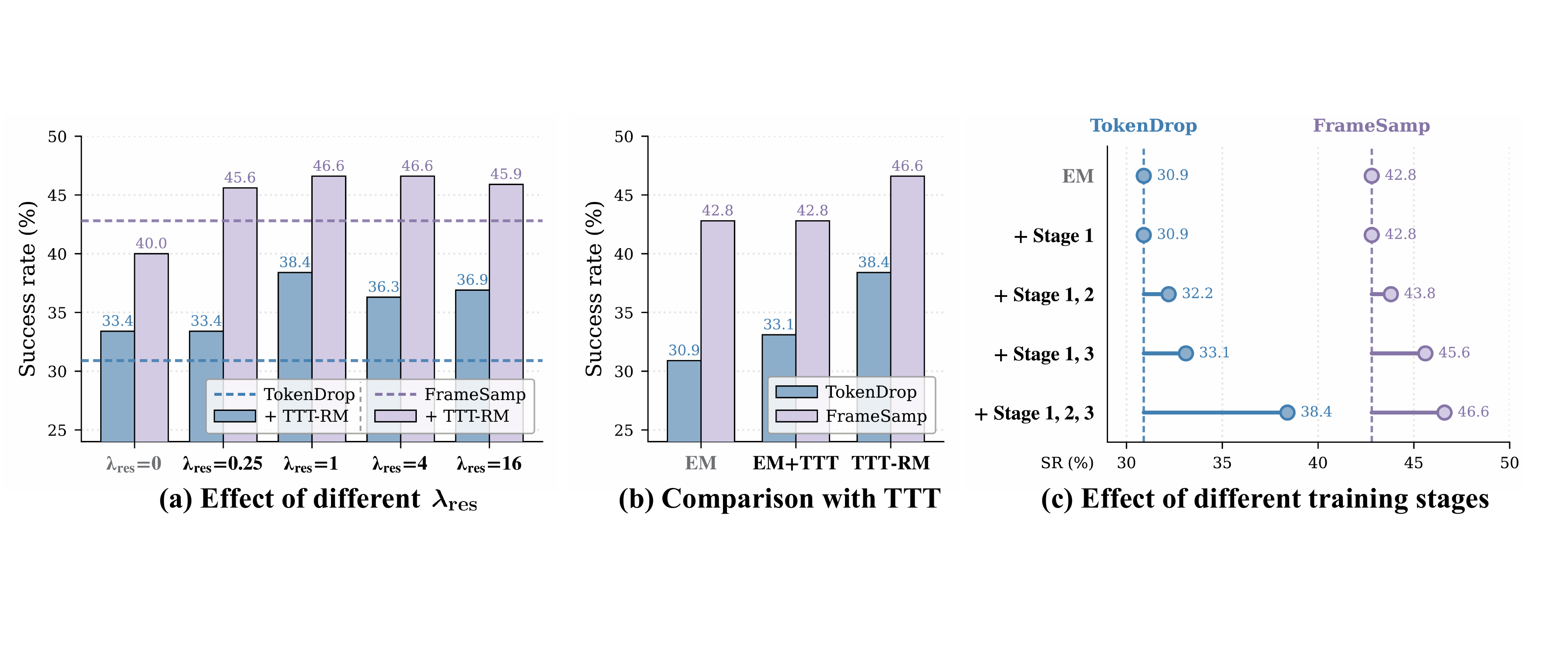}
    }
    \vspace{-12pt}
    \caption{\textbf{Ablation studies.} \textbf{(a)} shows the effect of the residual-loss weight $\lambda_\mathrm{res}$. Disabling residual supervision ($\lambda_\mathrm{res}=0$) can degrade performance, while moderate values yield the strongest results. \textbf{(b)} compares TTT-RM with a naive EM+TTT combination, showing that simply adding TTT to an EM-equipped policy provides limited benefit. \textbf{(c)} examines the effect of our three-stage training pipeline, with the full procedure achieving the best performance across both EM variants.}
    \label{fig:ablations}
\end{figure}

\noindent\textbf{Ablations and analysis.} We conduct targeted ablations to isolate individual design choices and examine how they affect performance.
\begin{itemize}[leftmargin=10pt,topsep=-3pt,itemsep=1pt,partopsep=1pt,parsep=1pt]
    \item \textbf{Ablation on $\lambda_{\mathrm{res}}$.} Fig.~\ref{fig:ablations}(a) studies the effect of $\lambda_{\mathrm{res}}$ across two Explicit Memory mechanisms. Without residual supervision ($\lambda_\mathrm{res}=0$), adding the TTT pathway has mixed effects, improving TokenDrop but degrading FrameSamp relative to their EM baselines. Moderate values of $\lambda_{\mathrm{res}}$ yield consistent gains for both mechanisms, whereas larger weights provide no further benefit and can slightly reduce performance. These results show that decoder-defined residual supervision is important for obtaining consistent gains from the added TTT pathway.
    \item \textbf{EM--TTT integration.} Fig.~\ref{fig:ablations}(b) compares TTT-RM with a plain TTT augmentation using the same architecture, data, and training schedule but without residual supervision. Simply adding TTT to EM yields limited or no improvement. 
    We also conducted an additional experiment, in which we started from standalone TTT and added either TokenDrop or FrameSamp under the \emph{Expert} integration. This improves the average success rate only marginally, from 17.5\% to 18.1\%. These results further indicate that simply combining the two memory mechanisms is insufficient. The gains of TTT-RM come from explicitly training the fast-weight state to preserve residual information not already captured by the policy context.
    \item \textbf{Training pipeline.} Fig.~\ref{fig:ablations}(c) ablates the three training stages described in Sec.~\ref{sec:implementation}. Stage 1 does not change rollout performance because it trains only the history decoder, leaving the policy itself unchanged. Training with either Stage 2 or Stage 3 improves performance, while their sequential combination performs best for both Explicit Memory mechanisms. Stage 2 uses the same objective as Stage 3 but updates only the newly introduced Residual Memory components while keeping the inherited policy frozen. We hypothesize that this warm-up reduces early optimization competition between newly initialized and pretrained components, enabling more effective joint fine-tuning in Stage 3 despite the modest standalone gain from Stage 2.
\end{itemize}

\subsection{Real-World Experiments}
\textbf{Setup.} We consider three real-world manipulation tasks: (1) \textit{Pen to drawer}, in which the robot moves exactly one of the two pens in the top cabinet into a drawer and then closes the drawer; (2) \textit{Pick $N$ pens}, in which the robot places $N\in\{1,2,3\}$ pens into a pen holder according to the language instruction; and (3) \textit{Stow objects}, an eight-stage long-horizon task in which the robot opens the top cabinet, transfers a toy and tea box into an open drawer, and closes the cabinet and drawer, requiring approximately three minutes for successful completion. We use a single-arm mobile manipulator~\citep{tidybot} and collect approximately one hour of demonstrations per task for policy training. The deployed policy operates at 20~Hz with a fixed action-execution horizon.

We compare against three baselines: a memory-free \textbf{Base} policy, an \textbf{Explicit Memory} variant, and a \textbf{Test-Time Training} variant, all initialized from the same $\pi_{0.5}$ backbone. For the real-world Explicit Memory baseline, each memory entry is a compact interaction record containing the observation, commanded action, and resulting outcome rather than visual features alone. For the TTT baseline, we re-implement RoboTTT~\citep{robottt} under the same training setup as TTT-RM. We evaluate each method over 20 trials per task under varied initial configurations. Per-trial performance is measured using a task-progress score that awards integer points for successfully completed task stages. For controlled comparison, all methods use synchronous policy execution. Additional task definitions, scoring criteria, and training details are provided in the Appendix.

\begin{figure}[t]
    \centering
    \includegraphics[width=1.0\linewidth, trim=50 520 35 50, clip]{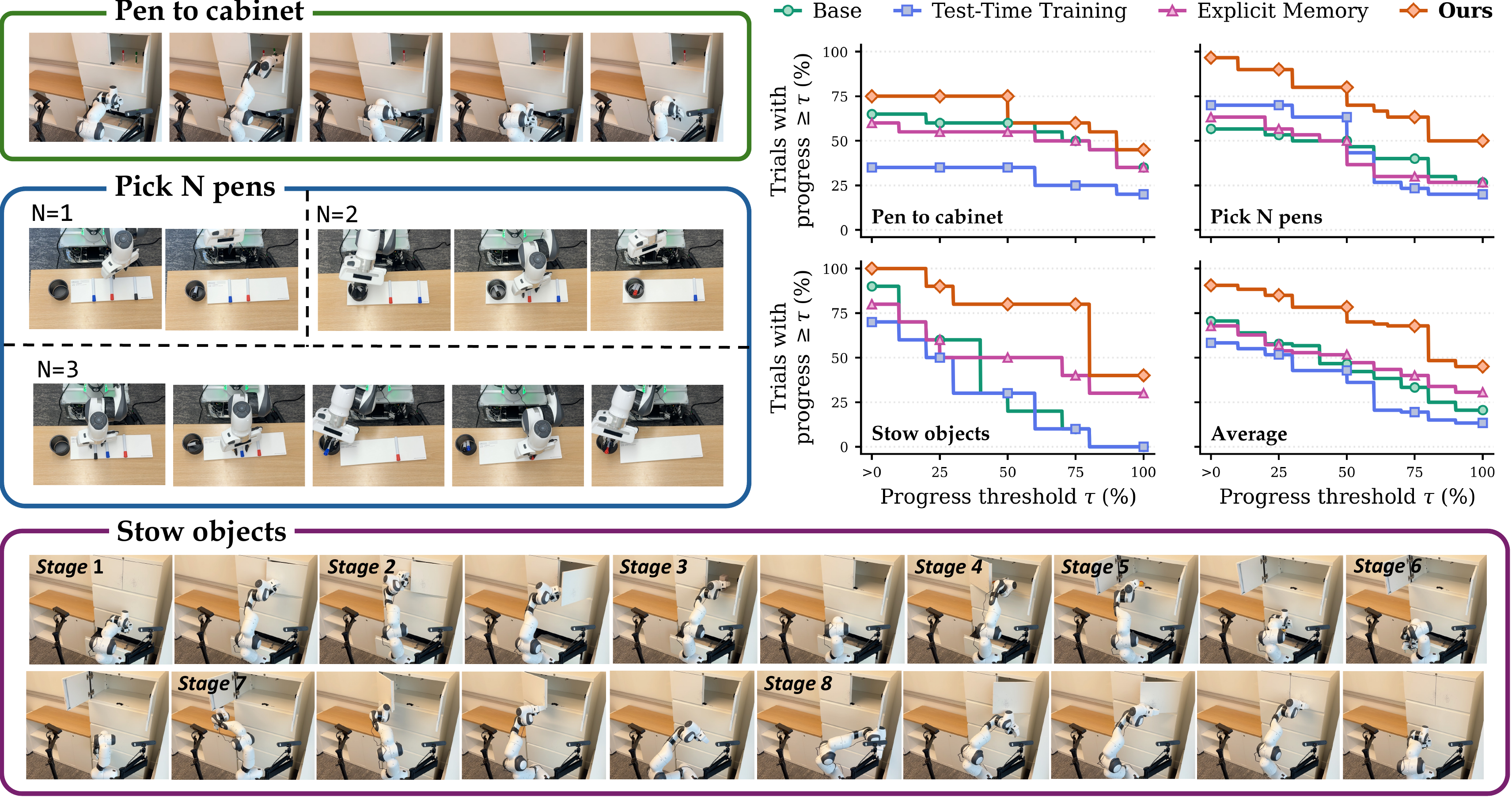}
    \vspace{-14pt}
    \captionsetup{
        justification=raggedright,
        singlelinecheck=false
    }
    \caption{\textbf{Real-world results.} Qualitative rollouts across the three manipulation tasks are paired with quantitative task progress profiles, which report the fraction of trials reaching or exceeding each progress threshold. Our method consistently achieves better performance across all task stages.}
    \label{fig:realworld}
\end{figure}

\noindent\textbf{Results.}
Fig.~\ref{fig:realworld} summarizes the qualitative and quantitative results across the three real-world tasks. For quantitative evaluation, we report the \emph{progress profile}, defined as the empirical survival function of per-trial task progress~\citep{progress_profile}. At each progress threshold, the profile measures the fraction of trials that reach or exceed that level of completion, providing a more detailed view of rollout quality than mean progress alone. TTT-RM consistently outperforms all baselines across all three tasks and at every progress threshold, indicating that its gains are distributed across different levels of task completion rather than driven by a small subset of successful trials. 
On \emph{Pen to drawer}, TTT-RM achieves a mean task progress of 65.5\%, compared with 55.5\% for the strongest baseline. The gains are substantially larger on the more memory-demanding tasks. On \emph{Pick $N$ pens}, mean task progress increases from 47.0\% for the strongest baseline to 73.5\%, and on \emph{Stow objects}, from 50.5\% to 77.0\%. In both cases, the improvement is also reflected throughout the progress profiles, with TTT-RM maintaining a larger fraction of successful rollouts as progressively later stages are considered. These larger gains suggest that Residual Memory becomes increasingly useful when successful execution requires tracking task progress over extended horizons.

Qualitatively, the memory-free Base policy exhibits characteristic failures under history-dependent ambiguity. The robot may continue acting after the required objective has already been achieved, oscillate between actions, or terminate prematurely when visually similar observations recur at different stages of execution. The TTT baseline performs comparably to, and in some cases worse than, the Base policy, suggesting that augmenting the policy with fast-weight memory does not by itself guarantee better history-dependent behavior and may interfere with action generation when the learned memory is not sufficiently informative. EM achieves stronger performance, with its advantage especially pronounced on the long-horizon \emph{Stow objects} task, where neither the Base nor TTT policy completes a full episode. TTT-RM further improves over EM by maintaining more consistent task progression across extended executions. In particular, it reduces repeated or premature actions at visually ambiguous stages and reaches later task stages that EM alone frequently fails to achieve.

\section{Conclusion} 
We introduced TTT-RM, which repurposes Test-Time Training as \emph{Residual Memory} to preserve task-relevant history not recoverable from the current observation and Explicit Memory. It first learns a history decoder to reconstruct historical representations from the policy's available context, then uses the reconstruction residual to supervise Residual Memory learning. The resulting representation is combined with retrieved memory to condition action generation.
TTT-RM generalizes across multiple memory designs, consistently outperforms diverse baselines, and supports sustained execution on a three-minute, eight-stage real-world stowing task.


\clearpage

\bibliographystyle{plainnat}
\bibliography{references}

\appendix
\renewcommand{\thefigure}{\thesection.\arabic{figure}}
\renewcommand{\thetable}{\thesection.\arabic{table}}

\end{document}